\documentclass[11pt]{article}

\usepackage[preprint]{acl}
\usepackage{preprint_header}
\usepackage{times}
\usepackage{latexsym}

\usepackage[T1]{fontenc}

\usepackage[utf8]{inputenc}

\usepackage{microtype}

\usepackage{inconsolata}

\usepackage{graphicx}

\usepackage{geometry}           
\usepackage{setspace}           
\usepackage{titlesec}           
\usepackage{hyperref}           
\usepackage{booktabs}
\usepackage{array}
\usepackage{amsmath}
\usepackage{tikz}
\usetikzlibrary{arrows.meta, positioning, calc, fit, backgrounds}
\usepackage{enumitem}
\usepackage{tabularx}
\usepackage{multirow}
\usepackage{tikz}
\usetikzlibrary{shapes.geometric, arrows.meta, positioning, calc, backgrounds}

\title{
Definitional Sensitivity in Media Bias Detection:\\ A Multi-Definition Dataset and Benchmark}

\author{
 \textbf{Martin Wessel\textsuperscript{1,2}},
  \textbf{Timo Spinde\textsuperscript{3}},
  \textbf{Jürgen Pfeffer\textsuperscript{1}},
  \textbf{Gianluca Demartini\textsuperscript{4}}\\
  \textsuperscript{1}Technical University of Munich,
  \textsuperscript{2}Center for Digital Technology and Management (CDTM),\\
  \textsuperscript{3}National Institute of Informatics (NII), Tokyo,
  \textsuperscript{4}University of Queensland\\
  \small{
    \textbf{Correspondence:} \href{mailto:m.wessel@media-bias-research.org}{m.wessel@media-bias-research.org}
  }
}

\begin{document}
\maketitle
\thispagestyle{preprintbox}

\begin{abstract}
Media bias detection relies on definitions and examples that specify what counts as bias, yet these specifications often vary across datasets or remain implicit, even when given the same name. 
Such variation makes it unclear whether models trained for the same bias category learn the same construct or different phenomena, a problem largely overlooked in prior work. 
We examine how definition choice affects bias annotation in a between-subjects experiment with 354 participants and a parallel evaluation with four LLMs. Participants and models rate six news articles across four bias categories using definitions that vary in conceptual framing and elaboration. 
Across 8,496 human and 28,800 LLM ratings, we find that the conceptual target of a definition drives annotation divergence, while construct-preserving elaboration does not: conceptual framing significantly shifts annotations for humans and does so even more strongly for LLMs. 
We discuss implications for construct specification in annotation protocols and prompt-based measurement, and consider how definitional sensitivity may propagate to downstream classification beyond media bias. 
We also release MUDD, the Multi-Definition Bias Detection Dataset.
\end{abstract}

\section{Introduction}
\begin{figure}[h!]
  \centering
  \includegraphics[width=\columnwidth]{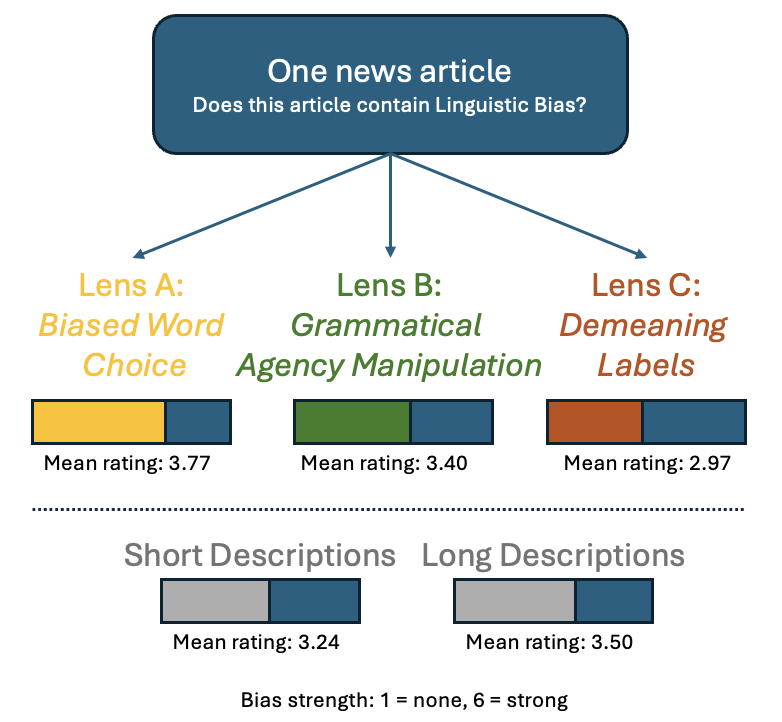}
  \caption{Same article, same bias category, three verdicts. Three standard conceptual definitions of \emph{linguistic bias} from the literature (biased word choice, grammatical agency manipulation, demeaning labels) yield significantly different mean bias ratings on the same 300 articles.}
  \label{fig:intro}
\end{figure}

Media bias, the slanting of media content through language, framing, or selection choices, can shape public opinion and political attitudes~\cite{entman2007framing, Spinde2025}.
Detecting it at scale relies on classifiers trained on human-annotated datasets, where annotators apply a given definition of bias to concrete texts.
That conceptual definition is the bridge between the abstract construct of ``bias'' and the labels that train and evaluate models.
Yet a review of 115 datasets finds that most rely on varying, often unstated, definitions of what counts as bias~\cite{WesselMBIB22}, and bias categories such as context bias or framing are overlapping and theoretically complex, leaving substantial interpretive room when a definition is applied to a full article.
Figure~\ref{fig:intro} illustrates the resulting risk: the same article, rated under three standard concepts of linguistic bias drawn from the literature, receives substantially divergent ratings.

If such definitional choices systematically push annotators toward different judgments on the same texts, then datasets that appear to measure the same construct may in fact capture different phenomena, undermining comparability across studies and the validity of classifiers trained on them.
Whether this happens, and how strongly, is an open empirical question: prior work documents that definitions vary across datasets but does not isolate the effect of that variation on annotation behavior.

We address this gap with a pre-registered, between-subjects experiment (a within-subjects design would let annotators compare definitions, confounding the effect itself).
In a $3 \times 2$ factorial design (concept $\times$ length), 354 U.S.-based participants each rate six of 300 news articles across four bias categories (linguistic, gender, political, and context bias) using one of six randomly assigned definitions per category.
The three concepts per category are theoretically grounded; the two lengths vary construct-preserving elaboration while holding the concept fixed.
We replicate the experiment with four large language models to compare human and LLM sensitivity to definitional variation.

Our results yield three contributions.
(1) We provide the first controlled evidence that \textit{what} a definition conceptually specifies, not \textit{how much} construct-preserving elaboration it provides, is a primary driver of annotation divergence.
We find significant concept effects across all four bias categories, robust to political orientation, gender attitudes, and clarity.
(2) The tested LLMs amplify this definitional sensitivity by approximately 1.3--4$\times$ in absolute rating differences relative to humans, and the direction of prompt effects can reverse across architectures.
(3) We release all 8{,}496 human annotations, 28{,}800 LLM annotations, 24~bias definitions, and full experimental materials as \textbf{MUDD}, the \textbf{Multi-Definition Bias Detection Dataset} for further research on definitional sensitivity in media bias detection.\footnote{Data, scripts and supplementary materials are available \href{https://github.com/Media-Bias-Group/Definitional-Sensitivity-in-Media-Bias-Detection}{here} and the preregistration \href{https://doi.org/10.17605/OSF.IO/X4Q2W}{here}.}

\section{Related Work}
\subsection{Definitional Variation Beyond Prompt Optimization}
Instruction and prompt wording are known to shift annotation and model behavior, but the effect of the underlying \emph{conceptual definition} of a target construct has received far less attention.
In media bias specifically, \citet{WesselMBIB22} find that the 115 datasets in the MBIB benchmark rely on inconsistent and often unstated definitions, a pattern they identify systematically.
\citet{spinde2023media} and \citet{rodrigo2024systematic} propose taxonomies that organize these conceptualizations, and \citet{Spinde2025TaxoMatic} argue, via an automated extraction framework, that the sheer volume and heterogeneity of available definitions makes their impact on downstream annotation a relevant concern.
In adjacent work, \citet{rottger2022two} show that descriptive versus prescriptive schemes yield fundamentally different hate speech datasets from the same content.
Definitional divergence is documented, but its causal effect on media bias annotation, a domain whose concepts are overlapping and theoretically complex, has not been tested directly.

\subsection{How Definitions Affect Human Annotators}
Within human annotation, guideline specificity is the closest proxy to what we study.
For media bias, \citet{fan_plain_2019} and \citet{hamborg_automated_2019} find that narrowly defined bias types yield more focused datasets, while \citet{salavati2024reducing} suggest clearer definitions improve quality at the cost of generalizability, and \citet{spinde_neural_2021} report persistent low inter-annotator agreement even with clear guidelines.
Beyond media bias, \citet{ross2017measuring} find that providing a hate speech definition does not improve agreement, and \citet{parmar2023don} show that guidelines at different detail levels produce significantly different labels, with more detail not necessarily helping.
Annotator identity adds a second source of variation: political and demographic characteristics predict subjective ratings~\cite{sap2022annotators, al2020identifying, talat2016you}, and annotator disagreement often reflects genuine ambiguity rather than noise~\cite{plank2022problem, uma2021learning}.

Together, these findings suggest definitions and annotator traits both move labels, but none isolates the causal effect of the definition itself while holding articles, interface, and annotators constant, motivating H1--H3:
\begin{description}
  \item[H1] Annotators assigned different conceptual definitions of the same bias category produce significantly different bias strength ratings for the same articles, and annotators assigned longer, more detailed definitions produce significantly different ratings than those assigned shorter definitions of the same concept.
  \item[H2] Annotators assigned different conceptual definitions or definition lengths report significantly different confidence in their bias judgments.
  \item[H3] After controlling for conceptual definition and definition length, the effects of annotators' political orientation and gender attitude on bias ratings are smaller than the concept effect.
\end{description}

\subsection{How Definitions Affect LLM Annotators}
LLMs increasingly serve as scalable annotation alternatives: \citet{horych2025promises} show LLM annotations can match or outperform expert data for media bias, and \citet{gilardi2023chatgpt} find ChatGPT outperforms crowd workers on several tasks at a fraction of the cost.
Their reliability, however, depends heavily on how the task is framed: \citet{pangakis2023automated} report substantial quality variation with task framing on subjective tasks, and \citet{tornberg2024best} show performance relies on prompt design and coding-instruction specificity.
What remains underexplored is sensitivity to the definition of the target construct itself, as opposed to surface-level prompt wording, which we test alongside the human experiment:
\begin{description}
  \item[H4] Definitional variation produces smaller variance in bias ratings among human annotators than among LLM annotations on the same articles and scale.
\end{description}

\section{Methodology}
\begin{figure}[h!]
  \centering
  \includegraphics[width=\columnwidth]{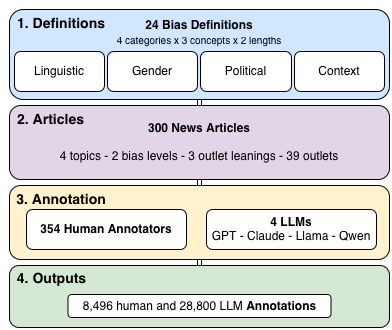}
  \caption{Experimental pipeline.}
  \label{fig:pipeline}
\end{figure}

\subsection{Research Design}

Our design aims to (1) isolate the effect of a bias definition from surface features such as length, register, and example count, (2) vary only theoretically separable concepts within a category so that any rating differences can be attributed to conceptual content, and (3) ensure that all definitions are comprehensible to non-expert annotators.
We run a pre-registered, between-subjects experiment using a $3 \times 2$ (concept $\times$ length) factorial design (Figure \ref{fig:pipeline} outlines the pipeline), applied independently across four bias categories: linguistic, gender, political, and text-level context bias.
We select these four to cover both mechanism-based bias (linguistic and context) and topic-based bias (gender and political), following the taxonomy of \cite{spinde2023media}.
For each category, we review existing bias definitions, classify each by its underlying mechanism, and select three concepts that target separable mechanisms and can independently vary within a single text, so that an article can exhibit one concept while being neutral on the others.
We fix the number of concepts to three \textit{a priori} to balance experimental power with feasibility in a between-subjects design.
We write the definition texts ourselves, based on the reviewed source conceptualizations, and standardize them across concepts for length, register, and example count, so that only the targeted mechanism varies.
We pilot-test the resulting definitions on Prolific ($n=24$) using the full main-study design supplemented with comprehensibility items; feedback from this pilot leads us to adopt easy-language phrasing throughout.
We provide a detailed description of the definition-derivation process, and final definition texts in Appendix \ref{app:definitions}.

We randomly assign each participant one of six definitions per category (three conceptual variants in either a short or long version) which use these to rate bias in news articles.
Short- and long-definition groups see only definitions of their assigned length, with no within-participant mixing.
For H3, we pre-specify a smallest effect of interest of $\pm 0.10$ scale points for covariate effects.
We operationalize length as construct-preserving elaboration: long definitions add explanation and examples that reinforce the same target mechanism.

Table~\ref{tab:subconcepts} summarises the three conceptual sub-definitions (C1--C3) per category.

\begin{table}[h]
\centering
\footnotesize
\setlength{\tabcolsep}{4pt}
\begin{tabular}{@{}lllll@{}}
\toprule
     & Linguistic & Gender & Political & Context \\
\midrule
C1 & Word choice & Visibility & Slant & Omission \\
C2 & Agency & Stereotypes & Coverage & Sources \\
C3 & Labels & Irrelevant focus & Framing & Emotion \\
\bottomrule
\end{tabular}
\caption{Concepts per bias category. For a detailed description and theoretical sources see Appendix~\ref{app:rationale}.}
\label{tab:subconcepts}
\end{table}

\subsection{Materials and Participants}

Bias perception varies systematically with article topic, outlet leaning, and framing style, so a diverse selection is necessary to avoid confounding definition effects with stimulus effects~\cite{Spinde2021e, spinde2023media}.
We therefore select 300 English-language news articles from \citet{newsmediabias_plus}, sampled across four topic domains (politics, social issues, economy, science) and stratified by NLP-predicted bias label (biased vs.\ unbiased, provided by \cite{newsmediabias_plus}) as well as source outlet political leaning (left, center, right), comprising 39 outlets spanning the ideological spectrum.

We run an a priori power analysis (power $= .80$, $\alpha = .05$, $f = 0.25$) which indicates that the most demanding test (a three-way interaction in the $3 \times 2 \times 3$ design) requires $N = 342$.
We recruit 354 participants through Prolific (not including 84 who drop out and 18 we reject for insufficient response time).
All participants are U.S.\ residents fluent in English, and we use Prolific's prescreening filters to approximate U.S.\ Census distributions for political orientation, gender, and age.

\subsection{Procedure and Measures}

Because both article characteristics and annotator characteristics shape bias judgments, we collect participant covariates alongside the annotation task~\cite{Spinde2021MBIC, sap2022annotators}.
We administer the study via Qualtrics.
After informed consent and a demographics questionnaire (age, gender, education, political orientation on an 8-point scale, two gender attitude items, news consumption habits), participants enter the annotation task.

Each participant reads six articles sequentially and rates each for bias across all four categories using their assigned definition.
Bias ratings use a 6-point Likert scale (1~=~Strongly disagree to 6~=~Strongly agree) in response to: \textit{``In my opinion, this article contains [category] bias.''}
Confidence is reported after each rating on a 7-point scale.
Definitions are accessible via a collapsible element on each page.
We embed an attention check that allows one retry; participants who fail twice are excluded.\footnote{The attention check tests whether participants grasp the general concept of bias after the definitions are displayed. The exact item is in the supplementary materials.}
After the task, participants rate the clarity of each definition using a six-item UEQ-adapted scale (covering understandability, value, supportiveness, quality, ease, and clarity, each on a 7-point semantic differential scale) and report overall reading attention.
We record page-level timing and definition re-click behavior throughout.
A full list of measures is provided in Appendix \ref{app:measures}.

We assign participants to conditions via a counterbalanced design: 27 covering sets of three definition profiles each, used twice (once per length group), ensure that all 300 articles are annotated under all six definitions across all four categories, yielding 7{,}200 article--category--definition combinations.

\subsection{LLM Annotation}

To ensure that the interpretation of definitions is directly comparable between humans and models, we replicate the human task across four LLMs chosen to span provider and cost tiers: two frontier commercial models, GPT-4o~\cite{openai2024gpt4o} and Claude Sonnet~4.6~\cite{anthropic2026claude}, and two same-size open-weight models from different providers, Llama~3.1~8B-Instruct~\cite{grattafiori2024llama} and Qwen2.5-7B-Instruct~\cite{qwen2.5}.
Each model receives a structured prompt containing the article text, one of the 24 definitions, and instructions to rate bias (1--6) and confidence (1--7).\footnote{The exact prompt template is available in the supplementary materials. We acknowledge that prompt form can influence LLM annotations; because all four models receive an identical prompt template that varies only in the injected definition text, any between-definition variation within a model is attributable to the definition itself rather than to scaffolding.}
Each model annotates all 300 articles under all 24 definitions ($7{,}200$ annotations per model) at temperature $0.0$, with each article--definition pair submitted as an independent query.
We access GPT-4o and Claude via commercial APIs and run Llama and Qwen via Hugging Face.
We analyse LLM annotations using the same statistical framework as the human data.

\section{Results}
We collect 354 completed responses (8{,}496 valid ratings) from U.S. Prolific workers: 174 women, 174 men, 6 non-binary; mean age 44.9 ($SD = 15.9$, range: 18--84); mean political orientation 2.99 ($SD = 2.28$) on a 0--7 scale (0~=~very liberal, 7~=~very conservative); 78\% consume news at least several times per week.
Educational attainment is diverse, with the largest groups holding a bachelor's degree ($n = 123$) or some college education ($n = 76$).
\footnote{Data quality was high: only 5 of 354 participants (1.4\%) failed an attention check, and none failed both. Self-reported reading attention was high ($M = 5.64$, $SD = 0.66$ on a 7-point scale), with no difference between length conditions. Bias ratings show no systematic drift across article positions ($r = .008$, $p = .43$), despite a 41\% decline in reading time from Article~1 (176.0\,s) to Article~6 (103.5\,s).}

\subsection{Concept Choice Dominates Definition Length}
\label{sec:concept_matters}

Category means range from $M{=}2.12$ (gender) to $M{=}3.61$ (context). Re-click rates are higher for linguistic (21.3\%) and context (21.0\%) than for gender (6.5\%) and political (6.5\%), suggesting definitional dependence is greatest where the construct is least intuitive. 
Because each participant is assigned a single definition per category, the concept and length factors are fully between-subjects and not crossed with participant; participant-level random intercepts are therefore largely unidentifiable from definition effects, and we model article as the primary source of non-independence.

\paragraph{Concept effects.}
One-way ANOVAs detect concept effects for all four bias categories.
Linguistic bias shows the strongest effect ($F(2,2115) = 48.57$, $p < .0001$, $\eta^2 = .044$), with all three concepts differing: Concept~1 (word choice, $M = 3.77$) $>$ Concept~2 (grammatical agency, $M = 3.40$) $>$ Concept~3 (demeaning labels, $M = 2.97$); all pairwise comparisons survive Bonferroni correction ($d = 0.25$--$0.53$).
Gender bias also shows a meaningful concept effect ($F(2,2115) = 11.64$, $p < .0001$, $\eta^2 = .011$), with Concept~1 (representational visibility, $M = 2.33$) rated higher than both Concept~2 (stereotypical traits, $M = 2.02$, $d = 0.22$) and Concept~3 (irrelevant focus, $M = 2.03$, $d = 0.22$), while Concepts~2 and~3 do not differ ($d = -0.001$, $p = .99$).
Context ($F(2,2115) = 3.97$, $p = .019$, $\eta^2 = .004$) and political bias ($F(2,2115) = 3.22$, $p = .040$, $\eta^2 = .003$) also produce statistically detectable concept effects, but with smaller effect sizes.

\paragraph{Length effects.}
Definition length, in contrast, has negligible effects.
Independent-samples $t$-tests reveal a length effect only for linguistic bias (long $M = 3.50$ vs.\ short $M = 3.24$; $t = -3.76$, $p = .0002$, $d = -0.16$); gender ($p = .27$), political ($p = .33$), and context bias ($p = .19$) show no length effects.
Two-way ANOVA confirms that for linguistic bias the concept effect ($\eta^2 = .043$) substantially exceeds both the length effect ($\eta^2 = .006$) and the Concept~$\times$~Length interaction ($\eta^2 = .005$).
Significant Concept~$\times$~Length interactions appear for linguistic bias ($F(2,2112) = 5.98$, $p = .003$) and, more weakly, for gender bias ($F(2,2112) = 2.83$, $p = .059$); political ($p = .272$) and context bias ($p = .537$) show none.

\paragraph{Mixed-effects confirmation.}
To account for repeated ratings of each article across definitions, we refit the main tests as linear mixed-effects models with a random intercept for article (\texttt{rating\,\textasciitilde\,concept\,*\,length\,+\,(1$\vert$article)}; likelihood-ratio tests). 
Concept remains significant in all four categories (linguistic $\chi^2(2)=110.0$; gender $\chi^2(2)=27.8$; political $\chi^2(2)=11.0$; context $\chi^2(2)=9.9$; all $p<.01$), while length is significant only for linguistic bias ($\chi^2(1)=14.8$, $p<.001$; gender, political, and context all $p>.09$), reproducing the ANOVA pattern. 
The article random intercept absorbs the expected between-article variance, most strongly for political bias ($\text{ICC}=.44$) and least for the more subjective categories ($\text{ICC}=.10$--$.17$), consistent with political bias being a comparatively shared construct (Section~\ref{sec:discussion}).

\paragraph{Annotator characteristics.}
ANCOVA models bound political orientation and gender attitude effects within $\pm 0.10$ scale points across all four categories (TOST, all $p < .05$), with definition variables accounting for virtually all explained variance. 
All 24 definitions receive comparable clarity ratings ($M \approx 3.84$--$4.07$), ruling out comprehension as a driver.

\subsection{Within-Definition Reliability}
\label{sec:iaa}

Although the between-subjects design assigns most article--definition combinations to a single annotator, we produce an overlap on 15\% of the 7{,}200 total combinations, each rated by 2--4 independent participants and balanced across the four categories.
This subset permits estimating \emph{within-definition} inter-annotator agreement.
Overall Krippendorff's $\alpha$ (interval) $= 0.39$ [95\% CI: 0.34, 0.44], with ICC(2,$k$) $= 0.54$, mean pairwise $|\Delta| = 1.36$ on the 1--6 scale, and 63\% of pairs agreeing within one scale point; $\alpha$ rises to 0.47 on the 200 combinations with $\geq 3$ raters, so the full-subset estimate is conservative.
Agreement varies sharply by category (Table~\ref{tab:iaa-category}).

\begin{table}[h]
\centering
\small
\begin{tabular}{lc}
\toprule
Category & $\alpha$ [95\% CI] \\
\midrule
Political & 0.56 [0.47, 0.63] \\
Context & 0.18 [0.07, 0.28] \\
Linguistic & 0.17 [0.06, 0.27] \\
Gender & 0.11 [$-$0.04, 0.24] \\
\bottomrule
\end{tabular}
\caption{Within-definition Krippendorff's $\alpha$ (interval) by bias category, computed on the multi-rater subset. 95\% confidence intervals from 1{,}000 combination-level bootstrap resamples.}
\label{tab:iaa-category}
\end{table}

\noindent Longer definitions yield slightly higher agreement ($\alpha = 0.40$ vs.\ $0.35$).
Given this low within-definition agreement, the concept effects reported in Section~\ref{sec:concept_matters} should be interpreted as aggregate shifts in the distribution of ratings across definitions, not as evidence that individual articles receive stable labels under any single definition; the systematic between-definition mean differences we observe arise in addition to, not instead of, substantial within-definition article-level variability.

\subsection{Confidence Reflects Decisiveness}

Confidence differs by concept for all four categories (all $p < .05$) but not by length (all $p > .20$). 
Crucially, confident annotators produce substantially more extreme ratings across all four categories ($r = .42$--$.45$, all $p < .001$), indicating that confidence indexes decisiveness rather than inter-annotator agreement.

\subsection{LLMs Amplify Definitional Sensitivity}
\label{sec:llms_amplify}
Table~\ref{tab:eta_comparison} compares concept effects across human annotators and four LLMs: GPT-4o, Claude Sonnet~4.6, Llama~3.1~8B, and Qwen2.5-7B, reporting $\eta^2$ from one-way ANOVAs alongside the largest pairwise absolute mean difference ($|\Delta M|$) and Cohen's~$d$.
Because $\eta^2$ denominators reflect different variance structures across multi-annotator and single-agent settings, $|\Delta M|$ and $d$ enable more direct cross-population comparison.

\begin{table}[t]
\centering
\footnotesize
\caption{Concept effects across three concept levels, by
annotator type and bias category: (a)~$\eta^2$ from
one-way ANOVAs, (b)~largest pairwise absolute mean
difference, (c)~largest pairwise Cohen's~$d$.}
\label{tab:eta_comparison}
\setlength{\tabcolsep}{4pt}
\begin{tabular}{@{}lccccc@{}}
\toprule
 & Human & GPT-4o & Claude & Llama & Qwen \\
\midrule
\multicolumn{6}{@{}l}{\textit{(a) Effect size ($\eta^2$)}} \\[2pt]
Linguistic & .044\rlap{***} & .171\rlap{***} & .165\rlap{***} & .142\rlap{***} & .346\rlap{***} \\
Gender     & .011\rlap{***} & .234\rlap{***} & .265\rlap{***} & .007\rlap{**}  & .104\rlap{***} \\
Political  & .003\rlap{*}   & .040\rlap{***} & .014\rlap{***} & .013\rlap{***} & .019\rlap{***} \\
Context    & .004\rlap{*}   & .045\rlap{***} & .075\rlap{***} & .040\rlap{***} & .150\rlap{***} \\
\midrule
\multicolumn{6}{@{}l}{\textit{(b) Max pairwise $|\Delta M|$}} \\[2pt]
Linguistic & 0.84 & 1.47 & 1.13 & 1.25 & 1.08 \\
Gender     & 0.34 & 0.96 & 0.80 & 0.06 & 0.59 \\
Political  & 0.19 & 0.76 & 0.40 & 0.50 & 0.37 \\
Context    & 0.21 & 0.73 & 0.78 & 0.74 & 0.71 \\
\midrule
\multicolumn{6}{@{}l}{\textit{(c) Max pairwise $|d|$}} \\[2pt]
Linguistic & 0.56 & 1.07 & 0.92 & 0.90 & 1.68 \\
Gender     & 0.25 & 1.09 & 1.33 & 0.16 & 0.80 \\
Political  & 0.12 & 0.50 & 0.28 & 0.27 & 0.35 \\
Context    & 0.15 & 0.53 & 0.75 & 0.46 & 0.85 \\
\bottomrule
\end{tabular}
\par\vspace{4pt}
\raggedright\scriptsize
\textit{Note.} Human $n{=}8{,}496$; LLM $n{=}7{,}200$ each.
{***}\,$p{<}.001$, {**}\,$p{<}.01$, {*}\,$p{<}.05$.
\end{table}

\paragraph{Amplification across categories.}
All four LLMs exhibit larger concept effects than humans across almost all categories (except Llama for gender).
Gender bias shows the largest amplification for GPT-4o and Claude (${\approx}2.4$--$2.8\times$ in $|\Delta M|$), context the most consistent across models (${\approx}3.4$--$3.7\times$); full magnitudes in Table~\ref{tab:eta_comparison}.
Category-level ordering is broadly preserved: linguistic produces the largest effects and political the smallest in every LLM (Spearman $\rho{=}0.80$ between Llama and Qwen on $\eta^2$).
Magnitude, however, is not architecture-invariant: Qwen dominates on linguistic and context bias, GPT-4o and Claude on gender, while Llama is least sensitive throughout, with its near-zero gender effect reflecting floor compression (all concept means $M{\approx}1.04$--$1.10$) rather than a general pattern.

\paragraph{Rating level and inter-model agreement.}
All LLMs rate below humans ($M{=}2.09$--$2.45$ vs.\ $3.11$); Llama's variance most closely matches humans, Qwen's is most compressed.
Per-article rating correlations are highest between GPT-4o and Claude ($r{=}.73$--$.82$) and moderate for human--LLM pairs ($r{=}.26$--$.51$).
Confidence scores are near-constant across all models ($SD{=}0.69$--$0.94$ vs.\ human $1.56$), so substantial rating shifts are unaccompanied by any signal of uncertainty.

\paragraph{Length effects and interactions.}
The concept-over-length pattern persists within LLMs but with model-specific signatures and inconsistent directionality: GPT-4o shows small length effects for gender and context ($d{=}0.12$, $p{<}.05$); Claude shows none (all $p{>}.09$); Llama shows effects for linguistic ($d{=}{-}0.09$) and context ($d{=}0.25$, the largest across annotator types); Qwen shows long\,$>$\,short effects for gender ($d{=}{-}0.19$) and context ($d{=}{-}0.26$).
Significant Concept\,$\times$\,Length interactions cluster on context bias; no LLM shows an interaction for political bias, whereas humans do for linguistic bias.

\paragraph{Article-level sensitivity.}
Article-level rating ranges across definitions span up to 5.0 on the 6-point scale. 
Mid-scale articles ($M{=}3.0$--$4.0$) show significantly higher definitional sensitivity than extremes for linguistic, gender, and political bias (human $d{=}0.29$--$1.00$; LLM $d{=}0.74$--$1.46$), but not for context ($d{=}{-}0.07$, $p{=}.57$). 
A temperature robustness check on GPT-4o confirms concept effects persist under non-zero model temperatures (Appendix~\ref{app:robustness}).

\section{Discussion}
\label{sec:discussion}

Across our four bias categories, we find that \emph{what} a definition specifies consistently affects human and LLM ratings of the same articles, while adding construct-preserving detail to a fixed target does not.

\subsection{Same Label, Different Constructs}
\paragraph{Conceptual framing, not elaboration, drives divergence (\textbf{H1}).}
Conceptual framing drives annotation divergence, while construct-preserving elaboration of a fixed target does not.
As we report in Section~\ref{sec:concept_matters}, concept effects are significant across all four bias categories, whereas length reaches significance only for linguistic bias and there with a far smaller effect size.

We interpret this as evidence that annotators form a mental working schema rapidly upon encountering a definition and then absorb additional elaboration into that schema rather than refining it.
Re-click rates are consistent with this, suggesting that definitional dependence is greatest where the construct is least intuitive.

\paragraph{Implications for construct validity.}
We interpret these results as indicating that, in our data, if two researchers operationalise ``media bias'' using different concepts within the same category, they likely produce systematically different datasets.
This happens not because their guidelines differ in specificity, but because they end up measuring different things, even when their wording implies the same category.

Our between-subjects design supports this interpretation.
Several plausible confounds are bounded: political orientation and gender attitude effects on ratings are statistically equivalent to zero within a $\pm 0.10$ bound after controlling for definition assignment (\textbf{H3}; all TOST $p < .05$), including the Concept $\times$ Gender Attitude interaction.
Within-definition agreement (Section~\ref{sec:iaa}) further clarifies this: political bias combines the highest within-definition $\alpha$ (0.56) with the smallest between-definition effect, suggesting a relatively shared, stable construct, while gender and linguistic bias show large definitional shifts despite low intrinsic agreement, marking them as both intrinsically subjective \emph{and} sensitive to how they are defined.
Absolute rating shifts underline this: for linguistic bias, a single definitional choice moves ratings by nearly a full scale point ($|\Delta M| = 0.84$ on a six-point scale; Table~\ref{tab:eta_comparison}).

\paragraph{Definitional choices matter most for borderline articles.}
Our article-level analysis indicates that definitional choices matter most where annotation is already difficult: for three of four categories, mid-scale articles show the largest rating shifts across definitions, while clear-cut cases are comparatively robust.
Researchers using different operationalisations of the same bias category are therefore most likely to diverge on borderline cases, which are also the cases classifiers most need to get right.

\paragraph{Confidence indexes decisiveness, not reliability (\textbf{H2}).}
Confidence varies significantly across concept conditions, but tracks rating extremity ($r = .42$--$.45$) rather than inter-annotator convergence.
Confident annotators commit more firmly to their individual reading without agreeing more closely with one another.
We therefore interpret confidence as indexing decisiveness rather than certainty, a distinction that matters for aggregation schemes that treat confidence as a reliability proxy.
All 24 definitions receive comparable clarity ratings ($M \approx 3.84$--$4.07$), with no short/long differences (all $p > .42$), so we can rule out comprehension difficulty as a driver.

\subsection{Humans Generalise, LLMs Follow the Prompt}
\paragraph{LLMs amplify concept effects across categories (\textbf{H4}).}
GPT-4o, Claude, and Qwen amplify the human concept effect across every bias category (Table~\ref{tab:eta_comparison}), and Llama does so in three of four categories.

We propose that the amplification factor reflects the semantic distinctiveness of concepts within a category.
For gender bias, our three definitions target semantically separable phenomena with limited lexical overlap, so an LLM prompted with one encounters few cues that would activate the others.
For linguistic bias, the concepts share surface features that may activate similar internal representations even under distinct prompts, compressing between-definition differences.
We suspect human annotators resist this pattern because they bring a schema-driven understanding of ``bias'' that partly activates neighbouring constructs regardless of which specific mechanism the definition targets, whereas LLMs lack this integrative prototype and map the provided wording more literally.
The Concept $\times$ Length interactions are consistent with this reading: LLMs consistently show significant interactions for context bias, whereas humans do not (Section~\ref{sec:llms_amplify}), pointing to different feature integration.

This literalism can help or hurt depending on the goal.
When the goal is to approximate the broader, everyday sense of ``bias,'' the same tight coupling to wording becomes a disadvantage: small changes in phrasing can shift, or reverse, the direction of the effect.

\paragraph{LLMs are not drop-in replacements for human annotators.}
All four LLMs rate bias systematically lower than humans and report near-constant confidence, so substantial rating shifts across definitions come with no signal of uncertainty. 
GPT-4o and Claude correlate strongly with each other ($r{=}.73$--$.82$) but only moderately with humans ($r{=}.26$--$.51$), which means cross-validating LLMs against one another can produce reassuringly high reliability that masks a shared departure from human judgment.
One might ask whether the observed effects depend on the specific concepts we select.
We sample from a larger definitional space rather than claiming taxonomic completeness, and the significant, ordered rating gradients indicate that annotators treat the concepts as distinct.
A GPT-4o robustness check at T=0.7 confirms concept effects persist under stochastic decoding (\ref{app:robustness}).

\subsection{Beyond Media Bias}
\label{sec:general_discussion}
Media bias is a domain where definitional fragmentation is already documented: existing taxonomies map the contested concepts~\cite{spinde2023media, rodrigo2024systematic, Spinde2025TaxoMatic}, and dataset heterogeneity~\cite{WesselMBIB22}.
Our results add a causal layer to this picture.
Prior work asks how to phrase bias questions for annotators and how to organise the field's taxonomies; we show that, even once the phrasing is fixed, the definition itself moves the labels.
Datasets such as MBIC~\cite{Spinde2021MBIC} and the 115 collections surveyed in MBIB~\cite{WesselMBIB22} are not directly comparable simply because they share a category label.

We expect the same logic to apply to other subjective labeling tasks whose target admits multiple theoretically grounded concepts.
A practical consequence for NLP is that, for such constructs, the choice of concept appears to matter more than the choice of surface phrasing, which shifts attention from prompt wording to construct specification in both human guidelines and LLM prompts.

\subsection{Practical Recommendations}
\label{sec:practical_recommendations}

We synthesise four recommendations from the analysis above.
First, \textbf{definitional reporting should become standard}: we recommend that researchers publish exact definition texts and state which conceptual tradition they draw from.
Second, \textbf{concept selection matters more than construct-preserving elaboration}: our null length effects imply diminishing returns from lengthening a definition once its target construct is fixed.
Third, \textbf{LLM-based annotation requires sensitivity testing}: we recommend annotating with at least two conceptually distinct definitions per category and reporting the variance, since sensitivity in one category does not generalise to others.
Fourth, \textbf{within media bias detection, definitions should be treated as experimental interventions, not neutral metadata}: the inconsistency across 115 media bias datasets~\cite{WesselMBIB22} reflects, a situation in which researchers are running different experiments under the same label.

\subsection{Future Work}
We see several natural extensions. Deliberately varying definition quality would test whether clarity moderates outcomes when free to vary. Because each participant rates each item once, our design cannot separate intra-annotator noise from definitional effects; embedding hidden duplicates or a test–retest subset would bound how much within-definition variability reflects response noise. Since sensitivity in one category does not predict another, a further step is testing whether concept effects transfer across categories, and whether the definition-dependent model rankings we observe make LLM leaderboards for subjective annotation definition-contingent. More broadly, our findings invite replication across other subjective NLP tasks, larger open- and closed-weight models, and multilingual settings.

\section{Conclusion}

Across 354 humans and four LLMs, we show that \emph{what} a bias definition specifies, not \emph{how much} construct-preserving detail it adds, drives annotation divergence, with LLMs amplifying this sensitivity architecture-dependently. We release the Multi-Definition Bias Detection Dataset (37,296 annotations across 24 definitions) to support treating definitions as experimental interventions.

\section*{Limitations}

Our experiment covers four bias categories (linguistic, gender, political, and text-level context bias) selected from a broader taxonomy~\cite{spinde2023media}; whether the observed definitional sensitivity generalizes to other bias types (e.g., racial bias, reporting-level bias, framing bias) remains an open question for future work. 
Our length manipulation is construct-preserving: long definitions elaborate a fixed target without introducing edge cases or negative examples that could redirect interpretation.
Varying the interpretation-redirecting scaffolding might risk blurring the concept/length distinction our design deliberately separates, a distinction already visible in annotation-guideline work outside media bias~\cite{parmar2023don, rottger2022two}.

Our sample consists of U.S.-based, English-speaking Prolific crowdworkers. Because bias perception varies with language, culture, and expertise~\cite{Spinde2021e, Spinde2021MBIC}, and bias measurement often fails to transfer cleanly to other languages, our findings should be read as concerning U.S. English-speaking annotators; replication with trained journalists, linguists, or non-U.S. participants may yield different absolute levels or amplification patterns. On the stimulus side, the NLP-predicted labels we use to stratify articles~\cite{newsmediabias_plus} are not ground-truth bias indicators and serve only to balance predicted-biased and predicted-unbiased items across our four topic domains (politics, social issues, economy, science), which exclude areas such as health or entertainment.
Each participant rates only six articles under a single definition, which strengthens causal identification through the between-subjects design but restricts reliability estimation to the 15\% multi-rater subset. 
Our three concepts per category are a small sample of the much larger definitional space~\cite{spinde2023media, rodrigo2024systematic, Spinde2025TaxoMatic, WesselMBIB22}; a broader or differently chosen set could plausibly produce larger effects.
We run the main LLM experiment at temperature 0.0 to maximise reproducibility~\cite{horych2025promises}, and our $T{=}0.7$ robustness check covers GPT-4o but not Claude, Llama, or Qwen. 
Results are bound to the four tested versions; larger models in either the open-weight (for example, 70B-class) or closed-weight tier remain untested, and prior work indicates that framing and prompt effects can interact with model capability in non-monotonic ways~\cite{pangakis2023automated, tornberg2024best}. 
We also do not manipulate the prompt scaffold itself (for example, chain-of-thought or role prompting); this keeps prompt form identical across conditions and annotator types but leaves open whether a different scaffold would compress or amplify the observed concept effects.
Finally, our covariates (political orientation, gender attitudes, news consumption, attention) are subject to social-desirability bias~\cite{Spinde2021e}. 
Reading times decline by 41\% across the session, from 176.0\,s on Article~1 to 103.5\,s on Article~6, which is consistent with fatigue; although we find no rating drift across article positions ($r = .008$, $p = .43$), we cannot rule out fatigue-by-definition interactions our design does not detect. Our study also captures one-shot annotation, and prior work on inter-annotator agreement suggests training can shift but not eliminate definition-dependent variation~\cite{spinde_neural_2021}.

\section*{Ethical Considerations}

The study is approved by the ethics review board of the University of Queensland.
Prior to participation, all annotators receive a detailed information and consent page describing the purpose of the study, the nature of the task, the data collected, the expected duration, and the compensation, and can only proceed after giving explicit informed consent.
Participation is fully voluntary, and participants can withdraw at any point without providing a reason.
We recruited participants through Prolific and compensated them at an average rate of \pounds14.34 per hour, well above Prolific's recommended fair-pay threshold.
We collect no personally identifying information beyond the platform-mediated pseudonymous IDs, which we remove before analysis and before release; all data released as part of the MUDD Dataset is fully anonymised.
The news articles used are drawn from an existing, publicly available dataset~\cite{newsmediabias_plus}, and we do not redistribute article text beyond what that source already makes available.
We intend the released dataset and materials to support research on definitional sensitivity and annotation methodology in subjective NLP tasks; we do not endorse their use for training classifiers that label outlets or individual journalists as biased in production settings, given the definitional sensitivity our results demonstrate.

\section*{Use of AI Assistants}

We use Claude Opus 4.8 and 4.7 (Anthropic) to polish author-written text, assist with presenting results, and improve the readability of our analysis and experiment scripts. All research contributions are the authors' own, and the authors reviewed and verified all AI-assisted outputs.

\section*{Acknowledgments}
This research was made possible through the TUM--UQ exchange program. We thank the ARC Training Centre for Information Resilience (CIRES) for kindly providing office space, and the lab of Prof.\ Gianluca Demartini for valuable discussions and feedback. 
We further thank the Center for Digital Technology and Management (CDTM) for making this work possible.

This work is partially supported by an Australian Research Council (ARC) Future Fellowship Project (Grant No. FT240100022).
This work was partially supported by JSPS KAKENHI Grant JP24H00732, by JST CREST Grants JPMJCR20D3 and JPMJCR2562 including AIP challenge program, and by JST K Program Grant JPMJKP24C2 Japan. It was also partially funded by the German Federal Ministry of Education and Research (BMBF) through the DAAD (German Academic Exchange Service).

\bibliography{main, custom, anthology}

\appendix

\section{Measures}
\label{app:measures}

\paragraph{Dependent variables.}
\begin{itemize}
    \item \textit{Bias rating:} 6-point Likert scale per article per category (1~=~Strongly disagree to 6~=~Strongly agree)
    \item \textit{Confidence:} 7-point scale per rating (1~=~Not at all confident to 7~=~Extremely confident)
\end{itemize}

\paragraph{Independent variables.}
\begin{itemize}
    \item \textit{Concept:} Three levels per category (between-subjects)
    \item \textit{Length:} Two levels: short vs.\ long definition (between-subjects)
\end{itemize}

\paragraph{Covariates and control measures.}
\begin{itemize}
    \item Political orientation (8-point continuous scale)
    \item Gender attitudes (two items)
    \item Definition clarity (six-item UEQ-adapted scale, 7-point semantic differential)
    \item Self-reported reading attention (7-point scale)
    \item Page-level reading time (seconds)
    \item Definition re-click frequency (count per article per category)
    \item Attention check pass/fail
\end{itemize}

\section{LLM Robustness Check}
\label{app:robustness}
To separate prompt sensitivity from sampling variance, we re-run GPT-4o on a stratified 60-article\,$\times$\,24-definition subset at $T{=}0.7$ with five samples per cell. 
Within-cell $SD$ ranges from $0.22$ (political) to $0.45$ (context), while the largest between-concept $|\Delta M|$ is $3.7\times$, $3.5\times$, $4.1\times$, and $1.3\times$ the within-cell $SD$ for linguistic, gender, political, and context bias. 
Concept effects remain significant at $T{=}0.7$ (linguistic $\eta^2{=}.230$, gender $.205$, political $.067$, context $.041$; all $p{<}.001$), matching or exceeding the $T{=}0$ results.

\section{Definitions}
\label{app:definitions}
\subsection{Definition Derivation Procedure}
\label{app:def_derivation}

The four bias categories are selected from a broader taxonomy of media bias types compiled from the literature \citep{WesselMBIB22, spinde2023media} along two dimensions.
As \textit{mechanism-based} categories, we select \textbf{linguistic bias} and \textbf{text-level context bias}, which capture complementary levels of textual organization: local linguistic choices versus global informational structure.
As \textit{topic-based} categories, we select \textbf{gender bias} and \textbf{political bias} for their prominence in the media bias literature.

\subsection{Selection of Concepts}
\label{app:subconcept_selection}

When designing the bias definitions, we have three goals. First, we want to isolate conceptual content from surface features such as length, register, and example count, so that any observed rating differences can be attributed to what a definition specifies rather than how it is phrased. 
Second, we want each definition to be rooted in an existing research tradition, so that the concepts we compare reflect genuine theoretical divides in the literature rather than variants we invented ourselves. 
Third, we want the concepts within a category to be analytically independent, so that a given article can in principle exhibit one while remaining neutral on the others, letting us interpret between-definition rating differences as effects of the definition rather than of inherent article properties.
To meet these goals, we follow a two-step process grounded in existing reviews of the media bias, communication, and computational linguistics literature~\citep{WesselMBIB22, spinde2023media, rodrigo-gines-etal-2023-unedmediabiasteam, Spinde2025TaxoMatic}. 
For each of the four categories, we first collect the conceptualisations of that category used in prior work and classify each by its underlying mechanism. In all four categories, this process yields at least three separable concepts which adhere to two selection criteria:

\begin{enumerate}
    \item \textbf{Theoretical grounding.} Each sub-concept is rooted in a distinct research tradition, ensuring that conceptual differences reflect genuine theoretical divides rather than surface-level variation.
    \item \textbf{Independent variability.} The three concepts target analytically distinguishable mechanisms such that a text can in principle exhibit one while remaining neutral on the others, and the three can vary independently within a single article without logical entailment.
\end{enumerate}

\noindent We note that these concepts sample from a larger definitional space rather than claiming taxonomic completeness. 
As discussed in Section~\ref{sec:practical_recommendations}, the significant and ordered rating gradients observed across just three concepts per category suggest that the full range of operationalizations used in the field would yield at least comparable heterogeneity.

\subsection{Definition Authoring}

The definition \textit{texts} shown to participants are author-constructed. 
This choice ensures that definitions within each category match in length, example count, register, and readability, isolating conceptual content as the sole variable. 
Each sub-concept is written in a short version (one to two sentences, one example) and a long version (extended explanation, three examples), with the long version adding elaboration and additional examples without altering the target construct. 
We deliberately optimise the three concept definitions within each category for diversity rather than minimal contrast: we choose examples and mechanism phrasings that sit as far apart as is defensible within the concept, so that any residual rating differences we observe are driven by the conceptual core rather than by a single contestable wording choice. 
If three separated operationalisations produce the divergences we report, small wording perturbations within any one of them are unlikely to erase the pattern, although they could shift absolute effect sizes.

We then subject all 24 resulting definitions to two validation steps. 
The authors review each definition for conceptual accuracy, length parity within a category, and register consistency. 
We then pilot-test the full set on Prolific ($n=24$) using the same study design as the main experiment, supplemented with open-ended comprehensibility questions after each category. 
The pilot confirms that participants can reliably distinguish between concepts within each category and leads us to adopt an easy-language register throughout.

We acknowledge that even with this process, the specific wording within each sub-concept represents one defensible realisation rather than the only one. Our claims concern between-sub-concept differences, not absolute rating levels.

\subsection{Rationale per category}
\label{app:rationale}
\subsubsection{Linguistic Bias}

For linguistic bias, we find three levels of analysis: semantic content, syntactic form, and social labeling. 
We select \textbf{biased word choice}~(C1), rooted in lexical framing and Entman's framing theory \citep{entman1993framing}, targeting how adjectives, verbs, and metaphors carry implicit valence;
\textbf{grammatical agency manipulation}~(C2), rooted in transitivity and responsibility attribution research \citep{henley1995syntax}, targeting how active/passive constructions obscure or highlight who acts; and \textbf{demeaning or stigmatizing labels}~(C3), rooted in labeling theory and dangerous speech research \citep{leader2016dangerous}, targeting how identity-linked terms marginalize groups.

\subsubsection{Gender Bias}

Gender bias definitions in the literature organize around three stages of representation. 
We select \textbf{representational visibility}~(C1), rooted in media representation and gender sourcing research \citep{macharia2020makes}, targeting systematic over- or under-representation in expert roles and quotes; \textbf{stereotypical trait attribution}~(C2), rooted in the Stereotype Content Model \citep{fiske2018model} and gender schema theory, targeting how identical behavior is framed differently by gender; and \textbf{irrelevant personal focus}~(C3), rooted in objectification theory \citep{fredrickson1997objectification}, targeting the foregrounding of appearance, family status, or likeability when irrelevant to professional role.

\subsubsection{Political Bias}

Political bias definitions map onto functions that operate at different stages of news production. 
We select \textbf{evaluative slant}~(C1), rooted in partisan slant and tone analysis research \citep{groseclose2005measure}, targeting explicit favorability through adjectives and verdict-like conclusions; \textbf{selective issue coverage}~(C2), rooted in agenda-setting theory \citep{mccombs1972agenda} and gatekeeping research, targeting which issues are spotlighted or suppressed for each side; and \textbf{interpretive framing}~(C3), rooted in framing theory \citep{entman1993framing} and interpretive journalism research \citep{iyengar1994anyone}, targeting how identical facts are cast as, e.g., ``success'' versus ``failure.''

\subsubsection{Text-Level Context Bias}

Context bias definitions cluster around modes of informational slant that correspond to omission, asymmetry, and addition. 
We select \textbf{omission of relevant facts}~(C1), rooted in information omission research \citep{entman2007framing} and journalism ethics, targeting facts a reader would need to interpret the central claim fairly; \textbf{asymmetric source treatment}~(C2), rooted in source balance and false balance research \citep{boykoff2004balance}, targeting whether opposing sides receive comparable space, quotes, and argumentative weight; and \textbf{gratuitous emotional detail}~(C3), rooted in emotional priming and irrelevant information research \citep{schwarz1991ease}, targeting background details included not for informational value but to steer the reader's emotional response.

\subsection{Full Definition Texts}

Tables~\ref{tab:def_linguistic}--\ref{tab:def_context} present all 24 bias definitions used in the experiment. Each of the four bias categories comprises three conceptual sub-definitions (C1--C3), each in a short version (S: one to two sentences, one example) and a long version (L: extended explanation, three examples). Definition texts are shown as presented to participants.

\begin{table*}[htbp]
\centering
\caption{Linguistic bias: complete definition texts.}
\label{tab:def_linguistic}
\footnotesize
\renewcommand{\arraystretch}{1.4}
\begin{tabularx}{\textwidth}{>{\raggedright\arraybackslash}p{2.2cm} c X}
\toprule
\textbf{Sub-concept} & \textbf{Ver.} & \textbf{Definition text} \\
\midrule

Biased word choice (C1)
& S
& Linguistic bias is when an article's word choice (adjectives, verbs,
  labels, metaphors) nudges the reader toward liking or disliking a
  person or group, even if the basic facts could be described
  neutrally.
  \newline\textit{Example:} Neutral: ``Police dispersed the crowd
  after midnight.'' Biased: ``Police crushed a mob after midnight.''
\\
\addlinespace

& L
& Linguistic bias is when an article's word choice (adjectives, verbs,
  labels, metaphors) nudges the reader toward liking or disliking a
  person or group, even if the basic facts could be described
  neutrally. This can happen through a single loaded word or through
  many small word choices that add up across the article, slowly
  tilting the tone in one direction.
  \newline\textit{Example~1:} Neutral: ``Police dispersed the crowd
  after midnight.'' Biased: ``Police crushed a mob after midnight.''
  \newline\textit{Example~2:} Neutral: ``The senator questioned the
  proposal.'' Biased: ``The senator attacked the proposal.''
  \newline\textit{Example~3:} Neutral: ``Protesters gathered outside
  the courthouse.'' Biased: ``Agitators swarmed the courthouse
  steps.''
\\
\midrule

Grammatical agency manipulation (C2)
& S
& Linguistic bias is when grammar (active vs.\ passive voice, who is
  named first, who is left unnamed) shifts responsibility or blame
  without changing the facts.
  \newline\textit{Example:} ``The minister cut disability benefits by
  8\%.'' (clear) vs.\ ``Disability benefits were cut by 8\%.''
  (decision-obscuring)
\\
\addlinespace

& L
& Linguistic bias is when grammar (active vs.\ passive voice, who is
  named first, who is left unnamed) shifts responsibility or blame
  without changing the facts. This can happen when an article
  consistently uses sentence structures that highlight certain people
  as decision-makers while making others seem like bystanders, even
  though both played an active role.
  \newline\textit{Example~1:} ``The minister cut disability benefits
  by 8\%.'' (clear) vs.\ ``Disability benefits were cut by 8\%.''
  (decision-obscuring)
  \newline\textit{Example~2:} ``The company fired 200 workers before
  the holiday.'' (agency clear) vs.\ ``200 jobs were lost before the
  holiday.'' (agency hidden)
  \newline\textit{Example~3:} ``Officers shot the suspect during the
  chase.'' (responsibility assigned) vs.\ ``The suspect was shot
  during the chase.'' (responsibility obscured)
\\
\midrule

Demeaning or stigmatizing labels (C3)
& S
& Linguistic bias is when an article uses demeaning, stigmatizing, or
  ``othering'' labels for people (nicknames, slurs, dehumanizing
  terms, mocking descriptors) instead of standard, respectful terms.
  \newline\textit{Example:} Biased: ``The neighborhood is being
  overrun by junkies.'' Neutral: ``The neighborhood has seen an
  increase in people with opioid addiction.''
\\
\addlinespace

& L
& Linguistic bias is when an article uses demeaning, stigmatizing, or
  ``othering'' labels for people (nicknames, slurs, dehumanizing
  terms, mocking descriptors) instead of standard, respectful terms.
  This also covers labels that reduce a person to one negative trait
  or that make certain groups sound like outsiders or threats,
  especially when a neutral alternative would work just as well.
  \newline\textit{Example~1:} Biased: ``The neighborhood is being
  overrun by junkies.'' Neutral: ``The neighborhood has seen an
  increase in people with opioid addiction.''
  \newline\textit{Example~2:} Biased: ``Illegals continue to flood
  across the border.'' Neutral: ``Undocumented immigrants continue to
  cross the border.''
  \newline\textit{Example~3:} Biased: ``The tree-huggers blocked the
  construction site again.'' Neutral: ``Environmental activists
  blocked the construction site again.''
\\
\bottomrule
\end{tabularx}
\end{table*}

\begin{table*}[htbp]
\centering
\caption{Gender bias: complete definition texts.}
\label{tab:def_gender}
\footnotesize
\renewcommand{\arraystretch}{1.4}
\begin{tabularx}{\textwidth}{>{\raggedright\arraybackslash}p{2.2cm} c X}
\toprule
\textbf{Sub-concept} & \textbf{Ver.} & \textbf{Definition text} \\
\midrule

Representational visibility (C1)
& S
& Gender bias is when people of one gender are systematically more
  visible in an article: quoted more, treated as experts more often,
  or centered as the main actors, while others are absent or relegated
  to minor roles.
  \newline\textit{Example:} The article quotes six ``experts'' (CEO,
  professor, analyst, doctor, spokesperson, union leader): all men,
  while women appear only as ``local residents'' with no expert role.
\\
\addlinespace

& L
& Gender bias is when people of one gender are systematically more
  visible in an article: quoted more, treated as experts more often,
  or centered as the main actors, while others are absent or relegated
  to minor roles. This is especially telling when the
  underrepresented gender is equally present and qualified in the
  field being covered, but editorial choices make it seem like
  expertise belongs mainly to one gender.
  \newline\textit{Example~1:} The article quotes six ``experts'' (CEO,
  professor, analyst, doctor, spokesperson, union leader): all men,
  while women appear only as ``local residents'' with no expert role.
  \newline\textit{Example~2:} A technology article interviews five
  male founders about industry trends while the only woman quoted is
  an office manager commenting on workplace culture.
  \newline\textit{Example~3:} A political analysis piece features
  four male commentators debating policy while the sole female voice
  is a voter sharing a personal story rather than analysis.
\\
\midrule

Stereotypical trait attribution (C2)
& S
& Gender bias is when the article assumes or reinforces gender
  stereotypes (who is ``naturally'' caring, emotional, tough, logical,
  suited to leadership, etc.)\ when describing people or evaluating
  their actions.
  \newline\textit{Example:} ``He delivered a firm, strategic
  response.'' vs.\ ``She responded emotionally and struggled to stay
  calm.'' (With no evidence that emotionality differed, just a
  gendered interpretive lens.)
\\
\addlinespace

& L
& Gender bias is when the article assumes or reinforces gender
  stereotypes (who is ``naturally'' caring, emotional, tough, logical,
  suited to leadership, etc.)\ when describing people or evaluating
  their actions. This includes cases where the same behavior is
  described differently depending on gender: for example, calling it
  ``confidence'' in one person and ``aggression'' in another.
  \newline\textit{Example~1:} ``He delivered a firm, strategic
  response.'' vs.\ ``She responded emotionally and struggled to stay
  calm.'' (With no evidence that emotionality differed, just a
  gendered interpretive lens.)
  \newline\textit{Example~2:} ``The male nurse showed unusual
  compassion for his patients.'' (implying compassion is unexpected
  for men) vs.\ ``The nurse showed compassion for her patients.''
  (implying it is natural for women)
  \newline\textit{Example~3:} ``She managed to lead the negotiation
  successfully.'' (implying surprise) vs.\ ``He led the negotiation
  successfully.'' (stated as expected)
\\
\midrule

Irrelevant personal focus (C3)
& S
& Gender bias is when the article focuses on appearance, sexuality,
  family status, or ``likeability'' in ways that distract from
  competence or actions when this focus would be unlikely for another
  gender in the same role.
  \newline\textit{Example:} ``The lawyer, wearing a tight red dress,
  arrived in court\ldots'' Her legal arguments are summarized in one
  line, while her look gets a full paragraph.
\\
\addlinespace

& L
& Gender bias is when the article focuses on appearance, sexuality,
  family status, or ``likeability'' in ways that distract from
  competence or actions when this focus would be unlikely for another
  gender in the same role. This means drawing attention to personal
  traits like looks or family life that have nothing to do with the
  person's job or achievements, especially when a person of another
  gender in the same position would not get that treatment.
  \newline\textit{Example~1:} ``The lawyer, wearing a tight red
  dress, arrived in court\ldots'' Her legal arguments are summarized
  in one line, while her look gets a full paragraph.
  \newline\textit{Example~2:} ``The CEO, a mother of three who still
  finds time to stay fit, announced the merger.'' A male CEO's family
  and fitness would rarely be mentioned in the same context.
  \newline\textit{Example~3:} ``The senator, known for her warm smile
  and approachable demeanor, introduced the defense bill.'' Her male
  colleague's bill introduction mentions only his policy credentials.
\\
\bottomrule
\end{tabularx}
\end{table*}

\begin{table*}[htbp]
\centering
\caption{Political bias: complete definition texts.}
\label{tab:def_political}
\footnotesize
\renewcommand{\arraystretch}{1.4}
\begin{tabularx}{\textwidth}{>{\raggedright\arraybackslash}p{2.2cm} c X}
\toprule
\textbf{Sub-concept} & \textbf{Ver.} & \textbf{Definition text} \\
\midrule

Evaluative slant (C1)
& S
& Political bias is when the article systematically favors or
  disfavors a political party or ideology through positive or negative
  evaluation (tone, adjectives, verdict-like conclusions), rather than
  sticking to balanced reporting.
  \newline\textit{Example:} ``The minister's plan is a bold fix that
  will finally restore order.'' vs.\ ``The opposition's proposal is a
  reckless gamble.''
\\
\addlinespace

& L
& Political bias is when the article systematically favors or
  disfavors a political party or ideology through positive or negative
  evaluation (tone, adjectives, verdict-like conclusions), rather than
  sticking to balanced reporting. This includes patterns where praise
  and criticism are handed out unevenly: one side's actions are
  consistently described with positive words while the other side's
  similar actions get negative ones.
  \newline\textit{Example~1:} ``The minister's plan is a bold fix
  that will finally restore order.'' vs.\ ``The opposition's proposal
  is a reckless gamble.''
  \newline\textit{Example~2:} ``The governor's swift action saved
  thousands of jobs.'' vs.\ ``The mayor's hasty decision put the
  budget at risk.'' (Both describe fast action, but with opposite
  evaluative framing.)
  \newline\textit{Example~3:} ``The party delivered on its promise to
  reform healthcare.'' vs.\ ``The party pushed through a
  controversial overhaul of healthcare.'' (Same policy outcome,
  different tone.)
\\
\midrule

Selective issue coverage (C2)
& S
& Political bias is when political coverage repeatedly highlights
  certain issues or scandals for one side while downplaying or
  ignoring comparable issues for another side, shaping what audiences
  think matters.
  \newline\textit{Example:} An article highlights Party~A's misuse of
  travel funds, but Party~B's similar audit finding is mentioned
  briefly in a late paragraph or not followed up.
\\
\addlinespace

& L
& Political bias is when political coverage repeatedly highlights
  certain issues or scandals for one side while downplaying or
  ignoring comparable issues for another side, shaping what audiences
  think matters. Within a single article, this can show up when one
  side's problems are described in detail while the other side's
  similar problems are mentioned only briefly or left out entirely.
  \newline\textit{Example~1:} An article highlights Party~A's misuse
  of travel funds in detail, but Party~B's similar audit finding is
  mentioned only in a single sentence in the last paragraph.
  \newline\textit{Example~2:} An article spends six paragraphs
  detailing one candidate's policy contradictions with quotes and
  timelines, while a rival candidate's comparable flip-flop is
  summarized in one line without evidence.
  \newline\textit{Example~3:} An article thoroughly documents
  lobbying ties of one party's officials with names and amounts, while
  the other party's similar connections are acknowledged only as
  ``both sides have faced scrutiny.''
\\
\midrule

Interpretive framing (C3)
& S
& Political bias is when the article frames political events so that
  readers are pushed toward one interpretation (e.g., ``success'' vs.\
  ``failure,'' ``defense'' vs.\ ``attack''), even if the same facts
  could support multiple reasonable readings.
  \newline\textit{Example:} ``The government finally caved and reduced
  the fee.'' (implies weakness) vs.\ ``The government responded to
  public concerns and reduced the fee.'' (implies responsiveness)
\\
\addlinespace

& L
& Political bias is when the article frames political events so that
  readers are pushed toward one interpretation (e.g., ``success'' vs.\
  ``failure,'' ``defense'' vs.\ ``attack''), even if the same facts
  could support multiple reasonable readings. This goes beyond single
  words to how the overall story is told: the order of information
  and the explanations chosen can make the reader lean toward one
  conclusion when other conclusions are just as reasonable.
  \newline\textit{Example~1:} ``The government finally caved and
  reduced the fee.'' (implies weakness) vs.\ ``The government
  responded to public concerns and reduced the fee.'' (implies
  responsiveness)
  \newline\textit{Example~2:} ``The president was forced to abandon
  the policy after mounting pressure.'' (implies defeat) vs.\ ``The
  president revised the policy in light of new evidence.'' (implies
  rational adaptation)
  \newline\textit{Example~3:} ``The opposition blocked the bill,
  plunging the legislature into gridlock.'' (implies obstruction)
  vs.\ ``The opposition exercised its oversight role by challenging
  the bill.'' (implies democratic function)
\\
\bottomrule
\end{tabularx}
\end{table*}

\begin{table*}[htbp]
\centering
\caption{Text-level context bias: complete definition texts.}
\label{tab:def_context}
\footnotesize
\renewcommand{\arraystretch}{1.4}
\begin{tabularx}{\textwidth}{>{\raggedright\arraybackslash}p{2.2cm} c X}
\toprule
\textbf{Sub-concept} & \textbf{Ver.} & \textbf{Definition text} \\
\midrule

Omission of relevant facts (C1)
& S
& Context bias is when an article leaves out important facts or
  constraints that a reasonable reader would need to interpret the
  main claim fairly (even if everything stated is technically true).
  \newline\textit{Example:} Article says: ``City spending on
  `administration' rose 20\% this year.'' Missing context: The budget
  category was renamed/reclassified, so the increase is mostly an
  accounting change, not new spending.
\\
\addlinespace

& L
& Context bias is when an article leaves out important facts or
  constraints that a reasonable reader would need to interpret the
  main claim fairly (even if everything stated is technically true).
  This also covers missing background like how the data was collected,
  historical context, or exceptions to a rule that would change how a
  reader judges the claim.
  \newline\textit{Example~1:} Article says: ``City spending on
  `administration' rose 20\% this year.'' Missing context: The budget
  category was renamed/reclassified, so the increase is mostly an
  accounting change, not new spending.
  \newline\textit{Example~2:} Article says: ``Crime in the district
  increased by 15\% this quarter.'' Missing context: The police
  department changed its reporting method to include incidents
  previously counted differently.
  \newline\textit{Example~3:} Article says: ``The drug was found to
  be effective in clinical trials.'' Missing context: The trial had
  only 30 participants, lasted two weeks, and was funded by the
  manufacturer.
\\
\midrule

Asymmetric source treatment (C2)
& S
& Context bias is when an article mostly presents one side's reasons,
  quotes, or evidence while giving the other side little space, weaker
  paraphrases, or no serious explanation, creating a tilted overall
  picture.
  \newline\textit{Example:} 8 paragraphs: detailed quotes from one
  side (``We have three reasons\ldots,'' data, anecdotes). 1~sentence
  for the other: ``Critics disagreed,'' with no reasons, no data, no
  quote.
\\
\addlinespace

& L
& Context bias is when an article mostly presents one side's reasons,
  quotes, or evidence while giving the other side little space, weaker
  paraphrases, or no serious explanation, creating a tilted overall
  picture. This imbalance shows not just in how much space each side
  gets but also in how their arguments are presented: one side gets
  direct quotes and named sources while the other is summed up in a
  single vague sentence.
  \newline\textit{Example~1:} 8 paragraphs: detailed quotes from one
  side (``We have three reasons\ldots,'' data, anecdotes). 1~sentence
  for the other: ``Critics disagreed,'' with no reasons, no data, no
  quote.
  \newline\textit{Example~2:} A healthcare article quotes three named
  doctors supporting a treatment with detailed explanations while
  opposition is summarized as ``some experts have raised concerns''
  without naming anyone or explaining what the concerns are.
  \newline\textit{Example~3:} A labor dispute article gives the union
  representative two full paragraphs of direct quotes with statistics
  while the company's response is paraphrased in a single sentence as
  ``the company denied the allegations.''
\\
\midrule

Gratuitous emotional detail (C3)
& S
& Context bias is when an article adds background details that are not
  necessary to understand the event but are likely to shape the
  reader's emotion or moral judgment (sympathy, anger, disgust),
  steering interpretation.
  \newline\textit{Example:} About a consumer complaint lawsuit, the
  article includes: ``The plaintiff has filed for bankruptcy twice and
  was once evicted,'' without explaining why this is relevant to the
  specific complaint.
\\
\addlinespace

& L
& Context bias is when an article adds background details that are not
  necessary to understand the event but are likely to shape the
  reader's emotion or moral judgment (sympathy, anger, disgust),
  steering interpretation. This includes personal history or past
  events that have nothing to do with the story but make the reader
  feel a certain way about the people involved, especially when such
  details are only added for some people but not others.
  \newline\textit{Example~1:} About a consumer complaint lawsuit, the
  article includes: ``The plaintiff has filed for bankruptcy twice and
  was once evicted,'' without explaining why this is relevant to the
  specific complaint.
  \newline\textit{Example~2:} In reporting a traffic accident, the
  article mentions: ``The driver, who had a prior conviction for petty
  theft ten years ago, collided with the barrier.'' The prior
  conviction has no relevance to the accident.
  \newline\textit{Example~3:} In covering a whistleblower's
  complaint, the article notes: ``The employee, who went through a
  contentious divorce last year, filed the report on Monday.'' The
  personal history has no bearing on the validity of the complaint.
\\
\bottomrule
\end{tabularx}
\end{table*}

\end{document}